# Selective Embedding for Deep Learning

Mert Sehri[1], Zehui Hua[1], Francisco de Assis Boldt[2], Patrick Dumond[1]

[1]*University of Ottawa, Ottawa, ON, K1N 6N5, Canada*

[2]*Instituto Federal do Espírito Santo, Vitoria, ES, 29173-087, Brazil*

**Abstract**

Deep learning has revolutionized many industries by enabling models to automatically learn complex patterns from raw data, reducing dependence on manual feature engineering. However, deep learning algorithms are sensitive to input data, and performance often deteriorates under nonstationary conditions and across dissimilar domains, especially when using time-domain data. Conventional single-channel or parallel multi-source data loading strategies either limit generalization or increase computational costs. This study introduces selective embedding, a novel data loading strategy, which alternates short segments of data from multiple sources within a single input channel. Drawing inspiration from cognitive psychology, selective embedding mimics human-like information processing to reduce model overfitting, enhance generalization, and improve computational efficiency. Validation is conducted using six time-domain datasets, demonstrating that the proposed method consistently achieves high classification accuracy across various deep learning architectures while significantly reducing training times. The approach proves particularly effective for complex systems with multiple data sources, offering a scalable and resource-efficient solution for real-world applications in healthcare, heavy machinery, marine, railway, and agriculture, where robustness and adaptability are critical.

## 1. Introduction

Deep learning is increasingly integrated into various facets of modern technology, influencing everything from personalized recommendations to speech recognition systems[1], and image processing[2]. Its capabilities are now embedded in many consumer products, such as smartphones, smart speakers, and driver assistance technologies[3,4]. By identifying objects, converting spoken words into text, and providing personalized content, deep learning has reshaped numerous industries, including healthcare[5], heavy machinery[6], marine[7], railway[8] and agriculture[9] (Figure 1).

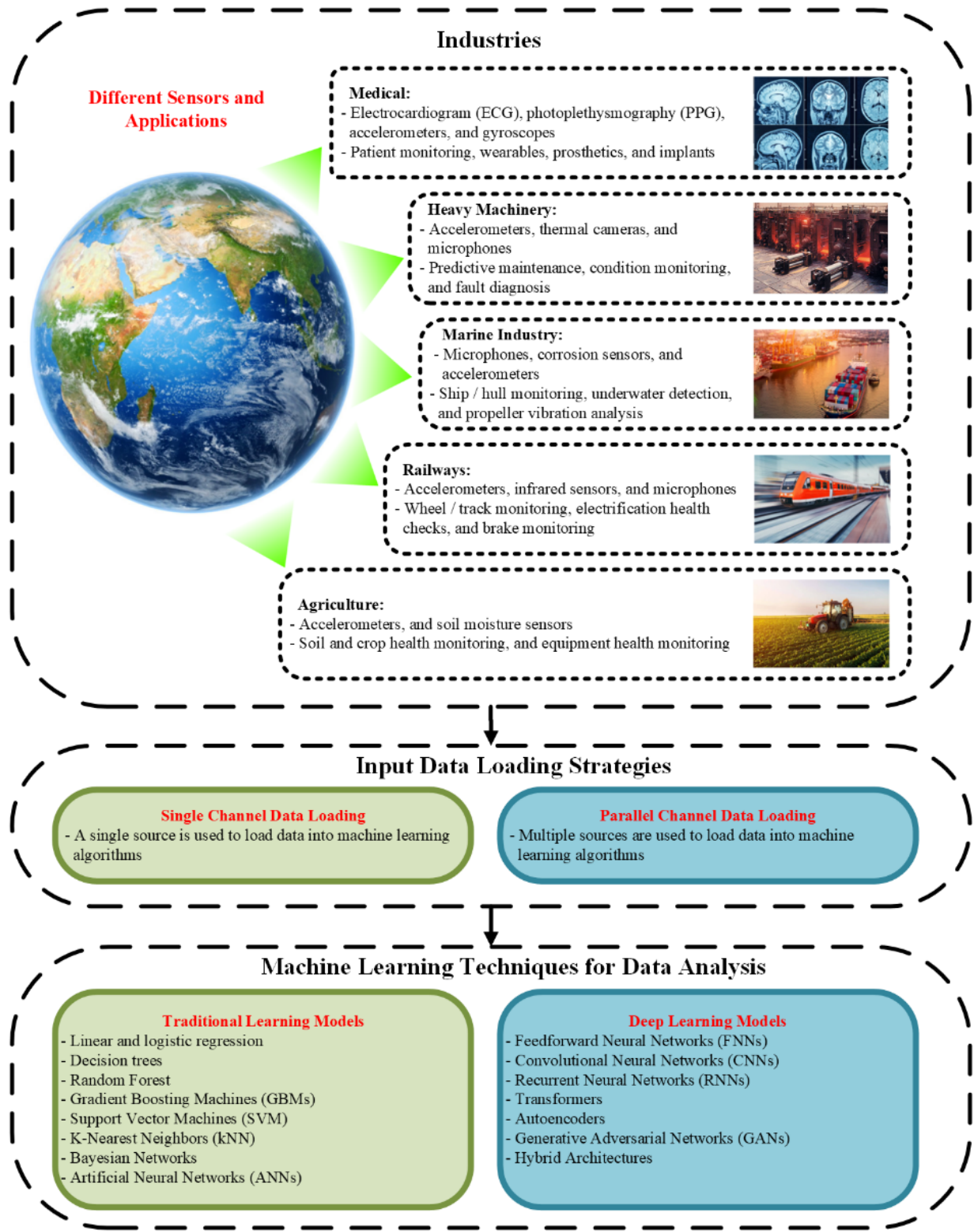

**Figure 1.** Overview of industries, data loading strategies, and machine learning techniques.

Unlike traditional machine learning methods that require manually crafted features and significant domain knowledge, deep learning offers a strategy for creating learning representations directly from raw data. This approach eliminates the dependency on feature engineering, enabling models to automatically extract important characteristics across multiple levels of abstraction. Starting from simple features, deep learning networks gradually build more sophisticated representations that prove to be critical for tasks such as classification and detection. However, the training of both traditional machine learning and deep learning models assumes that training and testing data share similar statistical patterns, which cannot always be guaranteed due to unknown domain shifts. As a result, training generalized models applicable to different instances of a similar application remains challenging. Moreover, the performance of these models may degrade significantly when given new data.

Nonetheless, advancements in deep learning help in tackling complex challenges in high-dimensional datasets, particularly those that conventional machine learning strategies struggle with. The success of deep learning spans across fields such as computer vision, speech recognition, genomics, and even drug discovery[4]. Additionally, the impact of deep learning in natural language tasks, ranging from sentiment analysis to language translation, further underscores its transformative potential.

Despite these achievements, there are still significant limitations when it comes to applying deep learning models to time domain data, especially when trying to generalize under nonstationary conditions or across different sources of data. Importantly, time domain data stems from highly diverse fields (Figure *1*) and plays an important role in understanding what is happening over time. Unlike areas such as computer vision and natural language processing, or even object detection, time domain data deals with diverse, dynamic, and often unpredictable environments. These complexities demand a robust approach to dataset preparation and model evaluation to ensure consistent performance across different scenarios that have not been sufficiently addressed in research, especially in terms of effectively loading datasets for optimal model generalization and efficiency.

For image classification tasks, high classification accuracy has been achieved when using large datasets such as ImageNet. The ImageNet dataset consists of many images for different categories of objects, where each image represents a different perspective of the same object. When loading an image, the model is improved when details are captured in three different channels known as red, green and blue (RGB). Combined with image normalization during dataset preparation, this method of image loading enables better generalization during the training of deep learning algorithms and helps reduce overfitting[10].

Historically, to train deep learning algorithms, standard methods like the holdout method have been widely used, where data is randomly divided into training, validation, and test sets. For instance, a commonly used data ratio includes 70% of the total data for training, 20% for validation and 10% for testing. While effective for some cases, the holdout method can be inadequate for smaller datasets or datasets lacking in diversity, where smaller datasets cannot provide sufficient data samples to train the model, which results in a model's poor ability to generalize to new data. To address this, *k*-fold cross-validation can provide new insight, where a dataset is split into *k* subsets, and the model is trained and validated *k* times, each time using a different subset as the validation set. This approach provides a more robust estimate of a model's performance and mitigates overfitting by ensuring that every data point is used for both training and validation, based on the assumption that training and validation datasets share the same feature distribution.

Enhancing the model's generalizability through innovative dataset structuring reduces overfitting and computational requirements, making it suitable for a variety of real-world scenarios. As deep learning technologies continue to evolve, these findings promise to support substantial advancements in classification analysis for deep learning applications. Nonetheless, conventional data handling techniques continue to struggle with domain shifts due to variability in machine behavior and operating conditions that degrade model performance[11]. In particular, dataset loading is often overlooked as a mechanism for improving generalization across environments even though inputs play a vital role in output model performance. To address this, a novel method for data loading is proposed based on the selective embedding of a diversity of data arranged into a single-channel input. The proposed method is specifically tailored to time domain data for enhancing generalization and improving model robustness under nonstationary conditions or dissimilar domains. A description of the proposed method compared to existing data loading methods is shown in Figure 2.

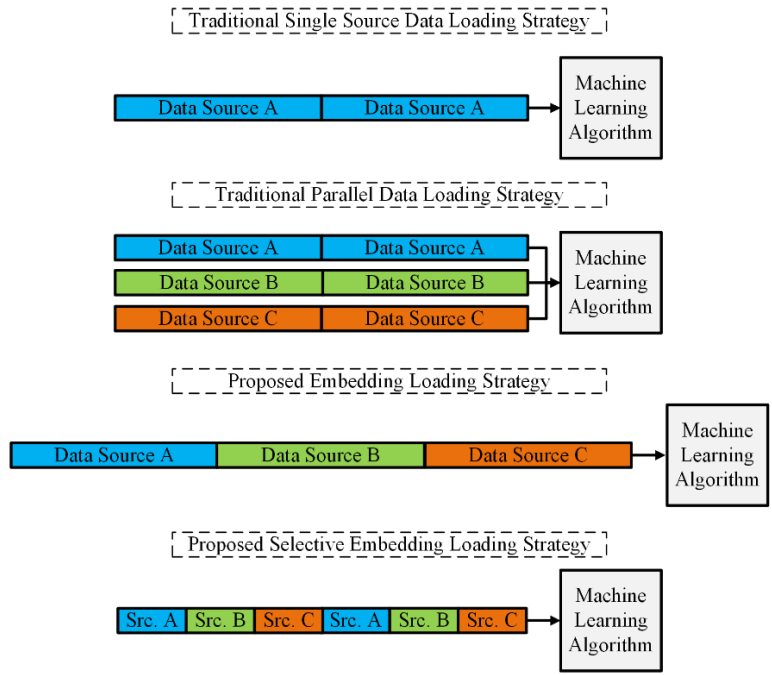

**Figure 2.** Existing and proposed data loading methods

## 1.1 Data Loading

In time domain analysis, data is often collected from either a single source or multiple sources and the content of this data can change over time due to nonstationary conditions. However, a common limitation in current machine learning model development is that training, validation, and test datasets are often split from limited field data of a single source, leading to poor generalization when new data is introduced for testing. Moreover, traditional approaches to loading multiple sources of data in parallel channels leads to long computation times and greater computational resource requirements, limiting the use of machine learning algorithms in the field. Therefore, a new approach to data loading is required to ensure better model training for nonstationary conditions or dissimilar domains.

To address this, a simple but effective data loading strategy is proposed that systematically loads data using selective embedding, which effectively alternates between loading short segments of data from different sources within a single channel. By selecting short segments of data from each source and then alternatively embedding these segments within a single input channel, model robustness can be enhanced, while reducing overfitting, and improving generalization without relying on complex algorithms. Furthermore, this can be achieved while reducing the computational requirements typically needed for such diversity in data.

## 1.2 Cognitive Theory

To better understand how data loading can be improved via selective embedding, artificial intelligence-based machine learning is explored in the context of human cognition and information processing. By looking at how information is received by the human brain, insight is gained to develop similar data loading strategies for machine learning algorithms. Here, data loaded into a machine learning algorithm is assumed to be like information processed by the brain. Thus, for the remainder of this paper, "information" and "data" will be used interchangeably.

Dual coding theory states that verbal and non-verbal information or data simultaneously work together to facilitate learning by creating better mental models[12]. The interconnectedness of this information is what the theory dictates as important for creating models that provide a richer representation of the system. When multiple sources of information or data are activated or loaded, learners have a better chance of understanding and retaining material[13]. In practice, the distribution of information from multi-source systems is traditionally represented by the parallel data loading strategy of Figure 2. For example, an industrial system can be represented by acoustic (verbal) data obtained from source A and vibration (non-verbal) data obtained from source B. In this case, both source A and B are monitoring the same piece of equipment providing a much richer picture of its state. A learning model is then created using data from both sources A and B loaded in parallel to understand its current or future state. Dual coding theory indicates that multi-source data providing information about the same system helps in creating better learning models than single-source data.

Cognitive load theory explains how the human brain manages information during learning and problem solving[14]. In essence, this theory implies that processes can be improved by reducing the total complexity of information. Unfortunately, multi-source data loaded in parallel is inherently more complex than single-source data. In machine learning, multitasking occurs when data is loaded in parallel, leading to small improvements in performance compared to traditional single channel loading at the expense of longer computation times when using a single processor or an increase in the computational resources required for multi-processor systems. Specifically, multitasking has been shown to overload the cognitive capacity of the brain increasing the time or effort it takes to complete tasks[15]. In the end, performance can be improved when using a parallel data loading strategy for multi-source systems at the expense of computational time or resources. Cognitive load theory indicates that embedding multi-source data within a single channel reduces complexity but does not account for the additional time required to process the increased amount of information. Therefore, an additional strategy is required.

Working memory is a fundamental concept in cognitive psychology that explores how human brains hold information or data temporarily and processes it to conduct mental tasks like reasoning, learning, and comprehension[16]. The theory breaks down information into important segments to prevent overloading of the working memory resulting in better retention, reasoning, and learning. If there is an overload of information or data in working memory, learning and problem solving become challenging and inefficient[17]. To accommodate for excessive amounts of information, the brain uses selective attention and filtering, where less relevant or redundant information is removed, keeping important information in order to solve the task at hand efficiently. Currently, excessive amounts of data or information collected by data scientists are often handled by manual or automatic preprocessing techniques to extract useful information before using it as an input. Instead, working memory theory indicates that machine learning could be improved by limiting data to smaller segments repeated in alternating cycles, rather than using all data from each source in sequence.

Drawing inspiration from the three theories presented, selective embedding considers the selective use of limited data embedded in an alternating single channel sequence to optimize the data loaded in machine learning algorithms. Short segments of information loaded in alternating sequences is inspired by the way the human brain is optimized for speed, energy efficiency, and adaptability in everyday decision-making. As illustrated in Figure 3, dual coding theory, cognitive load theory, and working memory theory form a conceptual map of how cognitive processes work together to inform learning. This integration offers a new perspective for developing more human-like deep learning data loading systems.

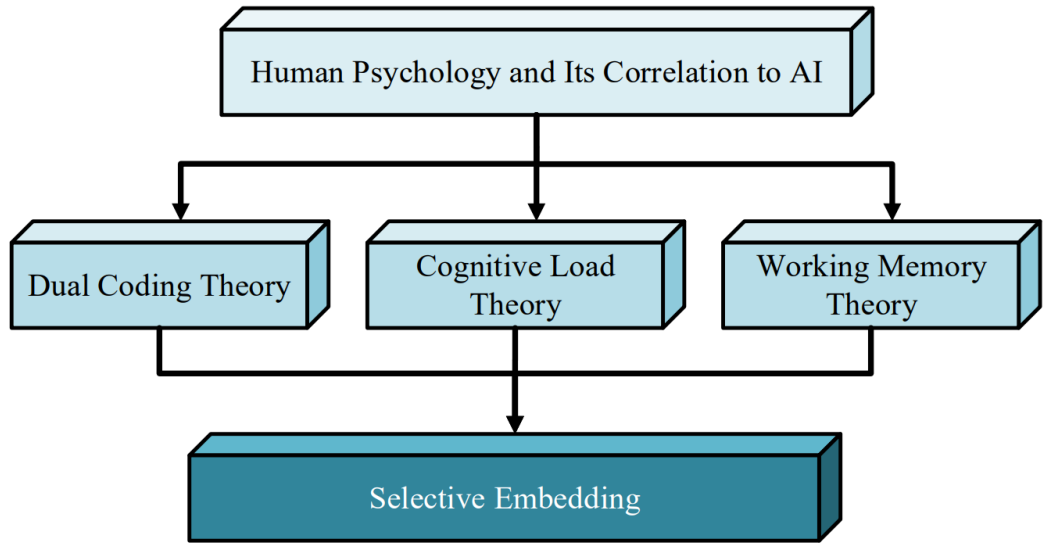

**Figure 3.** Basis of selective embedding

## 2. Methodology

Current data loading methods are limited (i.e., researchers typically preprocess data, then select a single-axis input or load multi-source data in parallel channels) for training deep learning algorithms. Single data source-based methods are fast, but trained models suffer from a lack of diversity in the learned features. On the other hand, multi-source data loaded in a single channel or via parallel channels increases test accuracies, but also the computation time or computational resources required to achieve this performance, respectively.

The proposed approach to data loading maintains or improves performance with a fewer number of diverse data when compared to traditional data loading methods, significantly improving computational efficiency. The current study demonstrates the improved efficiency of the method by running experiments on several publicly available datasets and shows the applicability of this data loading strategy for different time domain-based scenarios.

The proposed input data loading strategy alternates between sources of data for all phases of model training, validation, and testing within a single channel input, rather than using the same source throughout or multiple channels. By alternating sources of data (Figure 4), the approach provides diverse perspectives on the data characteristics of the signals of interest, allowing the model to learn from a broader range of data, reducing the risk of overfitting.

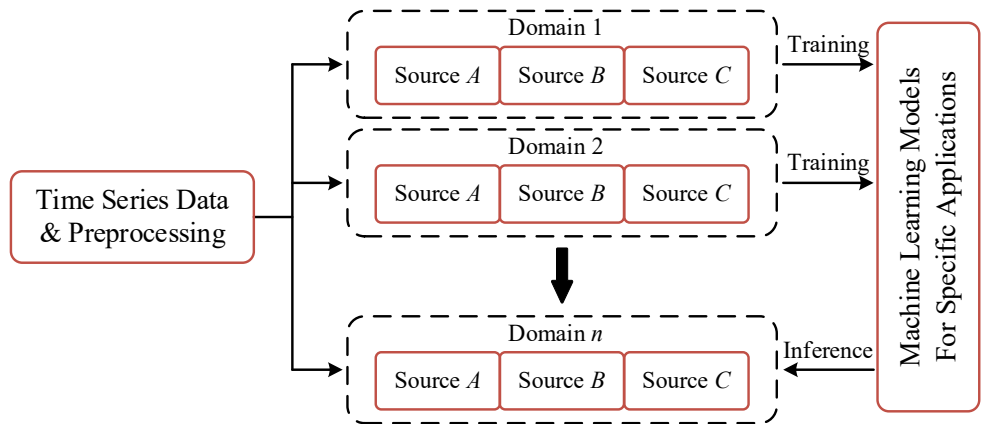

**Figure 4.** Selective embedding input data loading across different sources.

All objects have a fundamental resonant frequency that can be used as a feature in deep learning algorithms. To demonstrate the potential of selective embedding, time domain data is converted to the frequency domain for all datasets used herein and analyzed using simple deep learning algorithms.

The Fourier spectrum is especially useful for feature extraction in frequency data classification, as it is robust against external noise and changes in environmental conditions. For example, in heavy machinery, variations in loads may affect the amplitudes of frequencies for different objects or object conditions, the characteristic fault frequencies of these machines and their frequency patterns remain constant. This consistency makes the Fourier spectrum a powerful tool for identifying any object that has a resonant frequency.

When each data source is loaded using selective embedding, the frequency components used for classification remain consistent while providing a wide range of amplitude variations across different conditions, allowing the algorithm to differentiate between features more effectively. Classification refers to a technique where the machine learning model categorizes a set of data points (class) based on features using a labeled dataset to learn relationships between features of classes. This single-channel multi-data source input strategy uniquely diversifies the data fed into the model, capturing the inherent variability in a system's behavior and improving overall performance while reducing computational requirements due to the smaller size of training data. This proposed method leverages the strength of selective embedding to capture frequency domain representations, ensuring that critical frequency-specific features, like those associated with mechanical faults, are effectively learned. However, the choice of dataset domains must be considered before processing the data. A domain refers to a set of data files labeled with different classes created to understand the features, model architecture, and interpretation of results. Domain selection is intended to enhance the generalizability of object classification. Domain splitting refers to the careful creation of domains where each domain consists of a dataset from a different instance to ensure that data fed into training, validation, and testing preserves diversity while preventing data leakage between domains. This structured division strengthens the learning process and contributes to the consistent and repeatable performance of the proposed method across datasets.

A breakdown of selective embedding is shown in Figure 5, indicating a detailed version of the input data as it is used for training, validation, and testing, with different alternating data sources within a domain. For the experiments in this study, each domain consists of a data source file emanating from a distinct sensor that is divided into an arbitrary set of 1024 input signals preprocessed using the Fast Fourier Transform (FFT) with no overlap. This method can be used for any time series sensor-based dataset. Figure 5 helps in understanding the proposed data loading method as it compares to traditional data loading strategies.

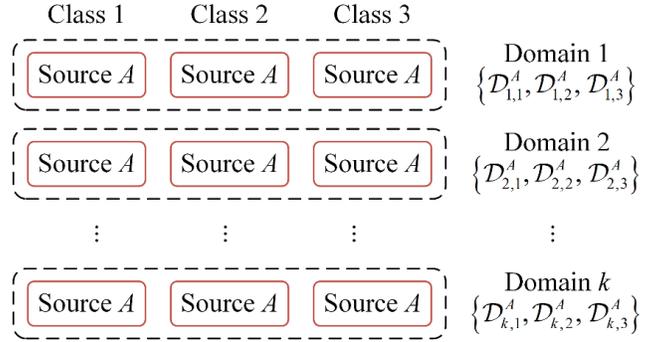

(a) Traditional loading (single)

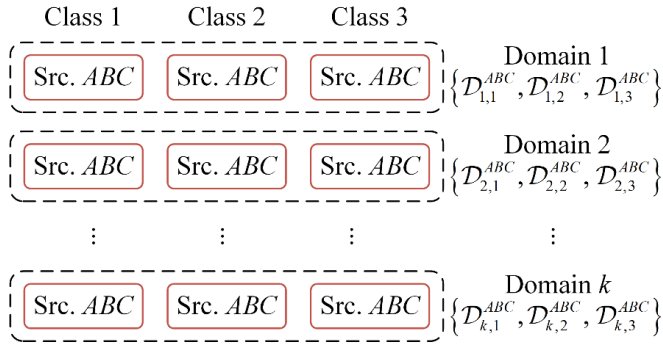

(b) Traditional loading (parallel)

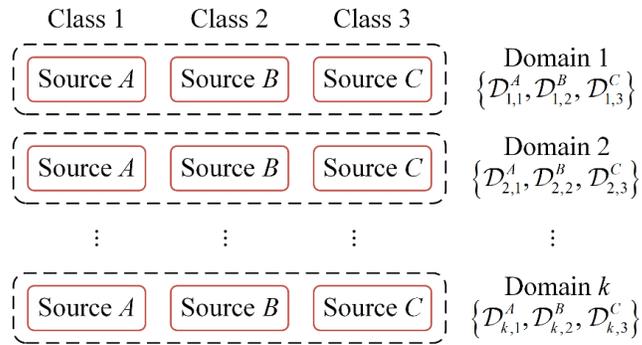

(c) Proposed loading (selective embedding)

**Figure 5.** Input data loading methods

**2.1 Data**

To demonstrate the use of selective embedding, heavy machinery and railway industry datasets are selected (Table 1). The first dataset is collected during steel manufacturing with the purpose of detecting the flow of steel and slag. The next three include bearing datasets (triaxial bearing vibration[18], CWRU[19], and UORED-VAFCLS[20,21]) collected from laboratory environments with seeded faults (triaxial: 0.9 mm, 1.1 mm, 1.3 mm, 1.5 mm, and 1.7 mm; CWRU: 0.007 in, 0.014 in, and 0.021 in) and naturally developed faults (UORED-VAFCLS). The fifth dataset contains data from an induction motor (UOEMD[22]) collected using acoustic and accelerometer sensors. The last dataset comes from the IEEE PHM Beijing 2024 Data Challenge provided by Beijing Jiaotong University[23], where data from a simulated train transmission system is analyzed. These datasets are selected to establish the superiority of selective embedding when compared to traditional methods.

**Table 1.** Datasets selected for this study.

| Dataset | Dataset Name | Machine Component | Specifications | Sampling Frequency | Industry | Sensor Type |
|---|---|---|---|---|---|---|
| Dataset 1 | Steel Slag Flow Dataset[24] | Flow Control Shaft | 16 different flow conditions | 6,400 Hz | Heavy Machinery | Triaxial Accelerometer |
| Dataset 2 | Triaxial Bearing Vibration Data[18,25] | Bearing | SKF 6204-2Z/C3 | 10,000 Hz | | Triaxial Accelerometer |
| Dataset 3 | CWRU[19] | Bearing | SKF 6205 | 12,000 Hz | | Drive End and Fan End Accelerometers |
| Dataset 4 | UORED-VAFCLS[20,21] | Bearing | NSK6203ZZ, FAFNIR 203KD | 42,000 Hz | | Accelerometer and Microphone |
| Dataset 5 | UOEMD[22,26] | Induction Electrical Motor | Marathon Electric D396 | 42,000 Hz | | Accelerometer and Microphone |
| Dataset 6 | IEEE PHM Beijing 2024 Data Challenge | Train Transmission System | 20 and 40 Hz, 0,+10,-10 kN loads | 64,000 Hz | Railway | Triaxial Accelerometers and Current Clamp |

The six datasets are divided into the domains described in Figure 6. These domains are chosen to reflect real-world scenarios.

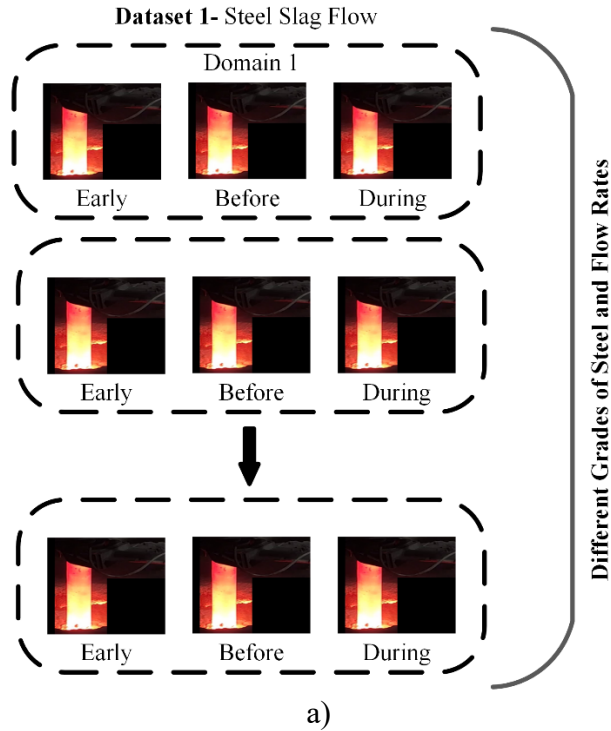

a)

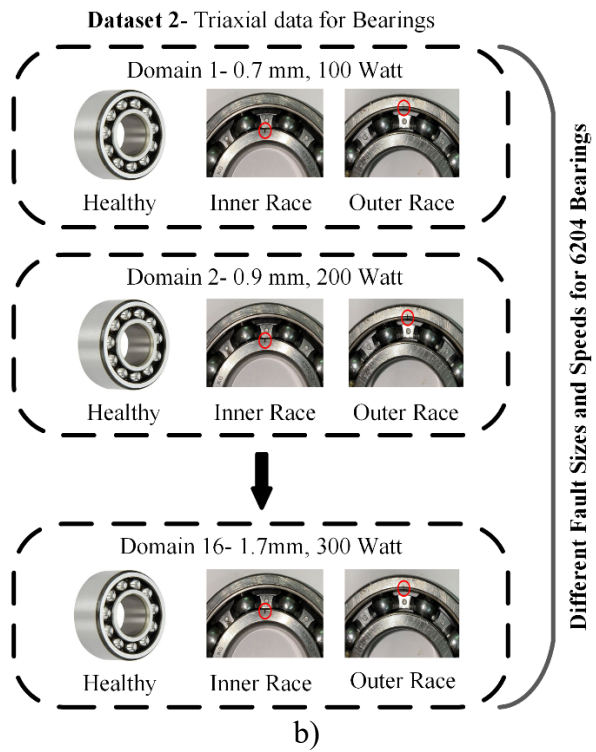

b)

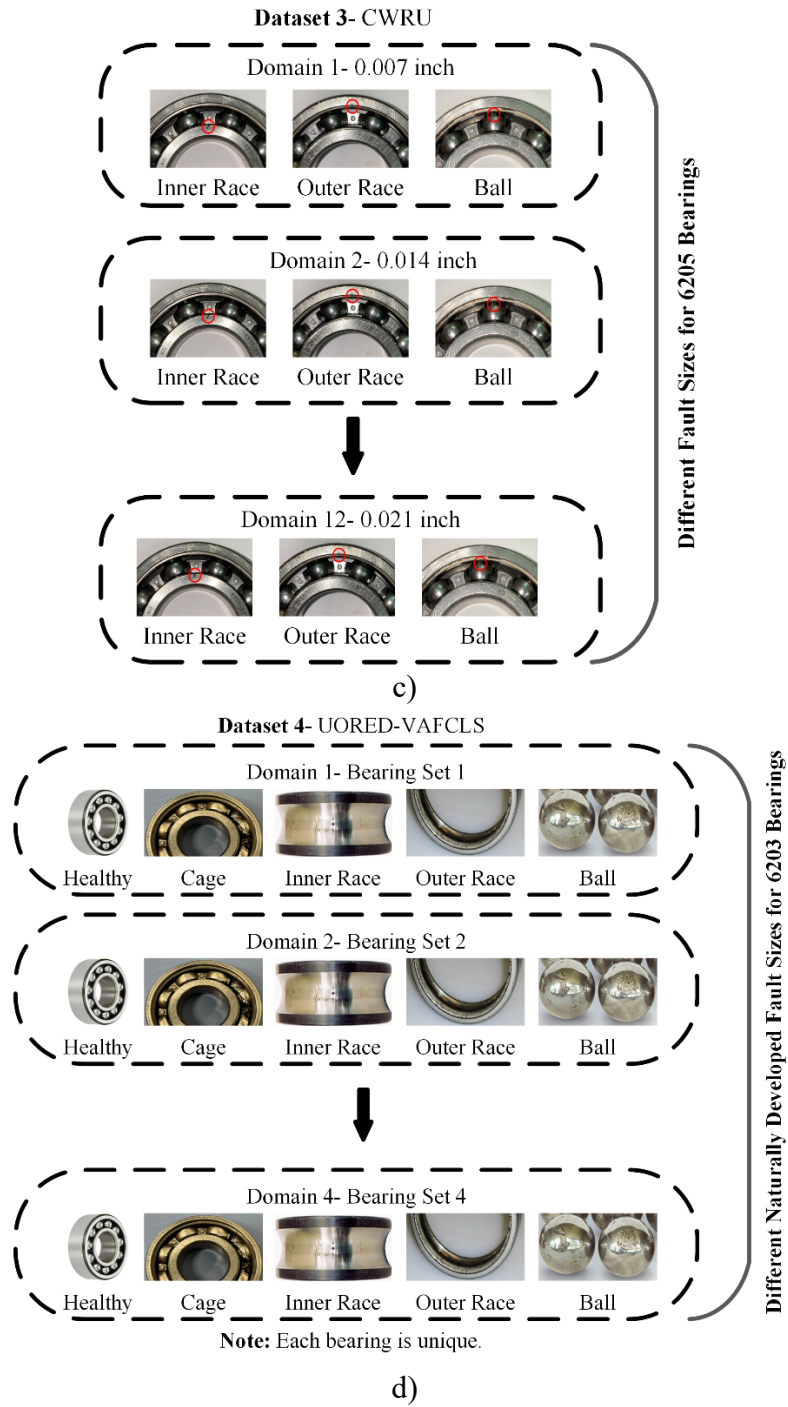

c)

d)

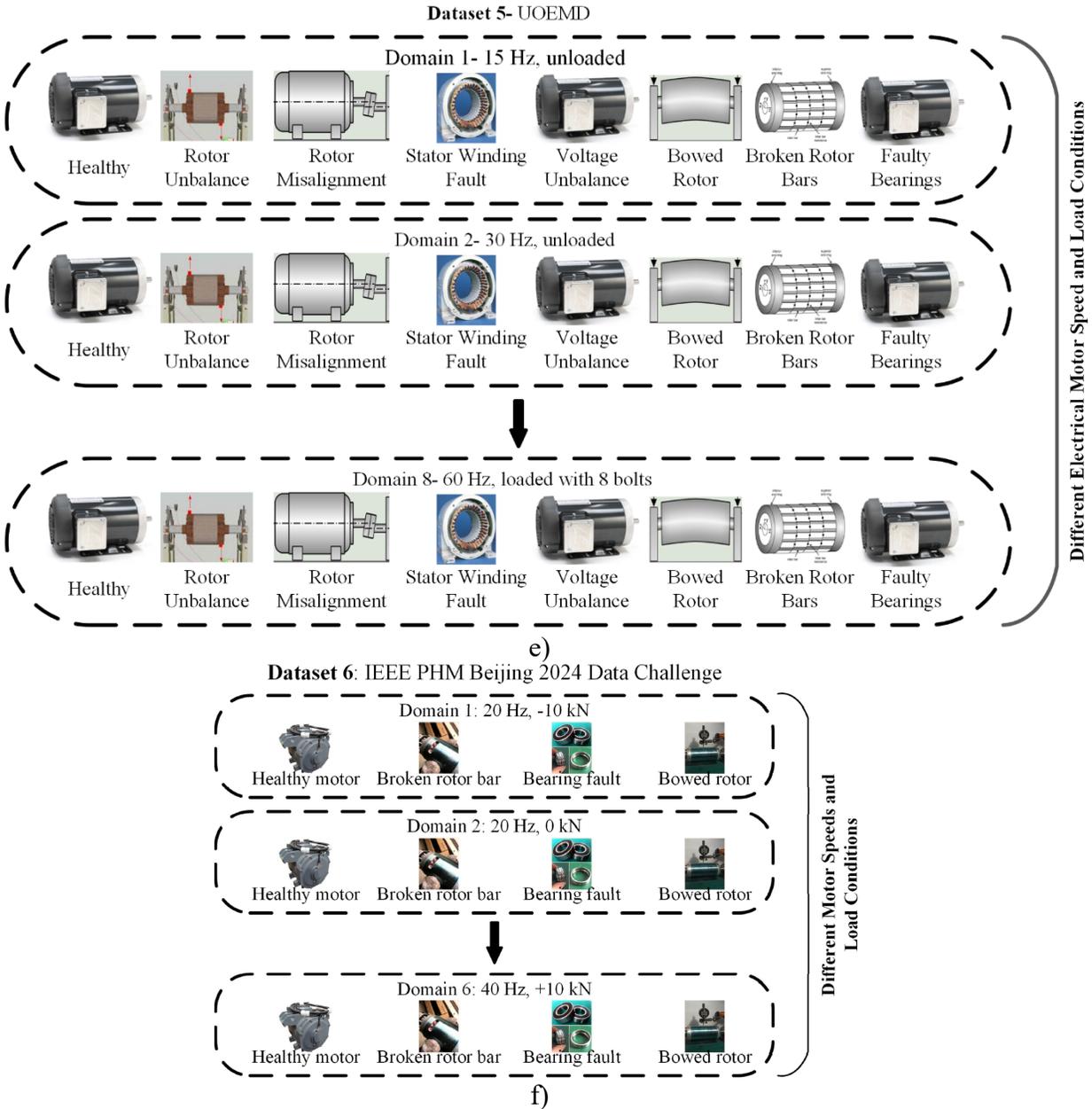

**Figure 6.** Domain loading of a) Dataset 1- Steel slag flow, b) Dataset 2- Triaxial dataset 6204 bearing, c) Dataset 3- CWRU 6205 bearing, d) Dataset 4- UORED-VAFCLS, e) Dataset 5- UOEMD, f) Dataset 6- IEEE PHM Beijing 2024 data challenge.

Dataset 3 (CWRU) is used in this section to demonstrate how time domain data is preprocessed using the FFT for training deep learning models. The time domain signals in Figure 7 a) illustrate vibration data across domains 0, 3, and 6 (inner race fault) for different fault severities. Even though the amplitudes vary between domains due to differences in fault sizes and sensor positions, the periodic nature of the signal remains evident. Important features for similar fault types also remain the same. Selective embedding helps distinguish these important similarities, improving the ability of deep learning models to generalize patterns by reducing input complexity but maintaining information diversity.

In the frequency domain (Figure 7 b)), the FFT-transformed signals show that the dominant frequency components across the domains match the fault frequencies expected for the operating conditions at hand, which is true for any object. However, the amplitudes at these frequencies differ between fault types, which provides critical information for fault classification. Other important features also remain similar. This consistent frequency response combined with amplitude variations ensures that selective embedding maintains robust fault characterization.

Selective embedding effectively leverages a reduced portion of the original dataset by eliminating portions of data from every source during data loading in an alternating sequence. For example, in the CWRU 12kHz bearing dataset, where each file contains 120,000 samples, the data is split into smaller, non-overlapping segments of 1024 FFT preprocessed samples coming from both the drive end (DE) and fan end (FE) sensors. These input segments are then alternated between DE and FE during training, validation, and testing. This allows for significant data reduction while retaining the critical features needed for condition classification, improving computational efficiency and maintaining a high accuracy performance.

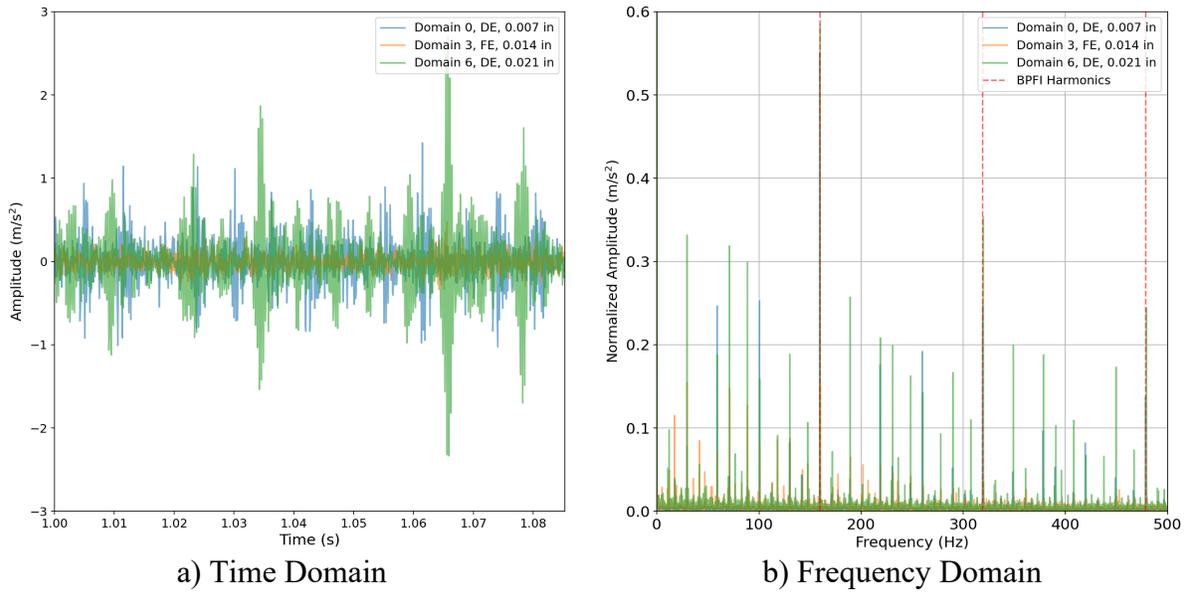

a) Time Domain   b) Frequency Domain

**Figure 7.** FFT class 1- inner race fault data domains 0, 3, and 6

Figure 8 further demonstrates selective embedding where time domain data segments used as inputs are highlighted in red boxes for both accelerometer and acoustic data from dataset 4 (UORED-VAFCLS). The numbering of each box shows how each data segment is loaded into the same channel of the deep learning algorithm for feature extraction. This can be done because data is collected simultaneously (i.e., at the same time) from each sensor, ensuring that frequency data from all sensors overlap, as shown in Figure 7.

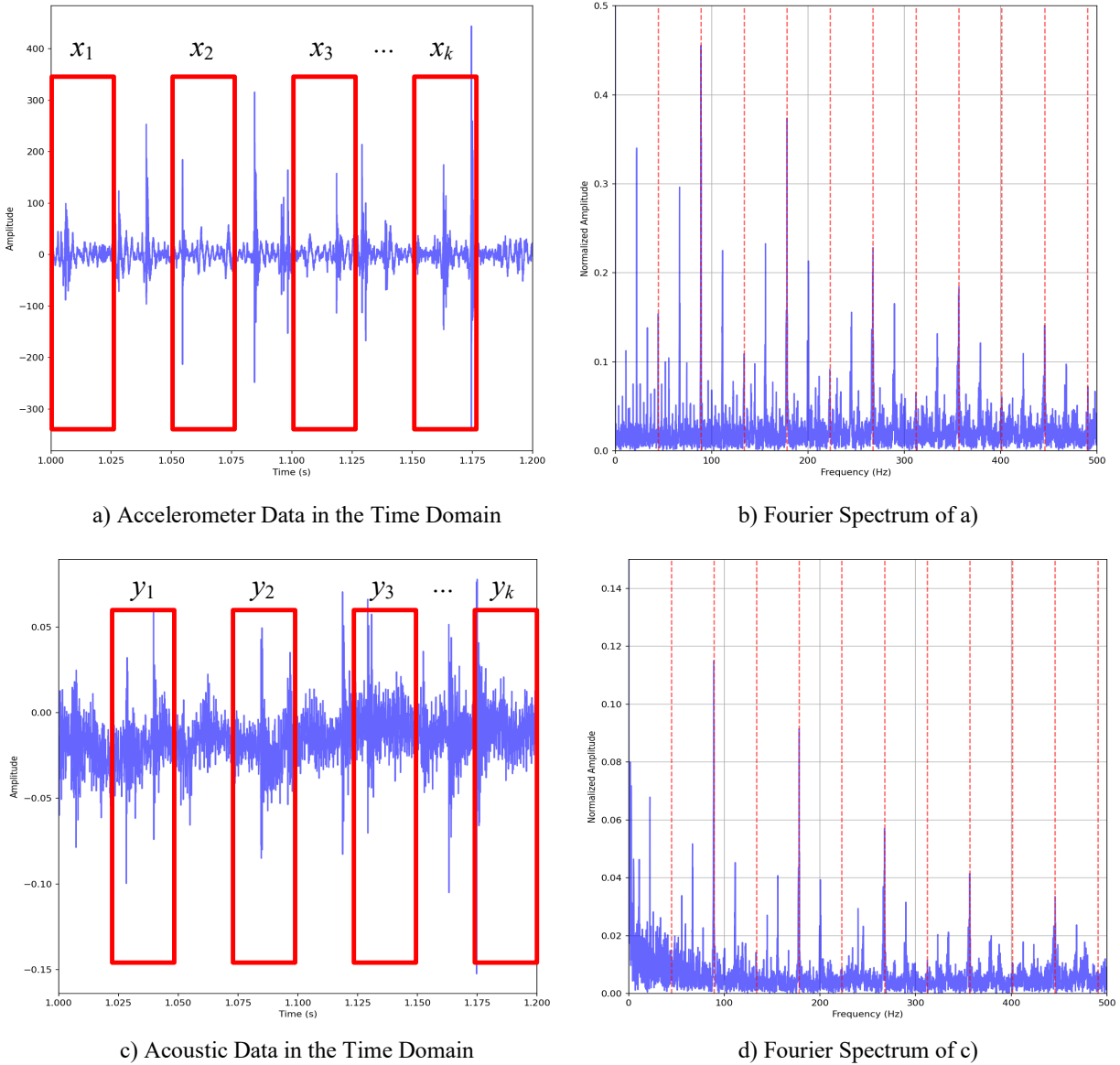

**Figure 8.** Demonstration of the selective embedding method for time domain and FFT data using dataset 4

Data contained within each red box of Figure 8 can be combined according to equation (1) to produce the data stream that is used as an input for the deep learning algorithm. Let values from the first data source $A$ (e.g., accelerometer in this example) which represent an arbitrary 1024 signal length from Figure 8 a) be represented by $x_1, x_2, x_3, ..., x_k$, where each sample is a feature vector (e.g., $x_1 = [x_{1,1}, x_{1,2}, x_{1,3}, and\ x_{1,1024}]$). Similarly, 1024 signal segments highlighted in red boxes from the other source $B$ (e.g., acoustic sensor), as shown in Figure 8 c), are represented by $y_1, y_2, y_3, \ldots, y_k$ (i.e., $y_2 = [y_{2,1}, y_{2,2}, y_{2,3}, and\ y_{2,1024}]$). $D$ represents the constructed dataset considering different class labels using the proposed selective embedding data loading approach, expressed as

$$D = \bigcup_{j=1}^{n_c} D_j \tag{1}$$

where for each class $j$ with a label of $l_j$, the data sample selected are formulated as

$$D_j = \begin{cases} \{(x_i^{(j)}, l_j) \mid i = 1, 2, 3, \ldots, k\}, & x_i^{(j)} \in A\ (j = 1, 3, 5, \ldots,\quad j \leq n_c) \\ \{(y_i^{(j)}, l_j) \mid i = 1, 2, 3, \ldots, k\}, & y_i^{(j)} \in B\ (j = 2, 4, 6, \ldots,\quad j \leq n_c) \end{cases} \tag{2}$$

where $x_i^{(j)}$ denotes the $i$-th sample from class $j$ in source $A$, $y_i^{(j)}$ denotes the $i$-th sample from class $j$ in source $B$ ($A$ denotes accelerometer data while $B$ denotes acoustic data).

## 3. Results and Discussion

Results for the six datasets included in this study are used as a benchmark when testing different deep learning architectures and indicate that selective embedding is effective and efficient with all networks. To demonstrate the ifmportance of selective embedding, a simple Convolutional Neural Network (CNN) and a complex CNN-transformer network are analyzed as examples in this section. Full results for all datasets using a CNN, as well as CNN-Long Short Term Memory (CNN-LSTM)[27], CNN-Gated Recurrent Unit (CNN-GRU)[28], Residual Network (ResNet)[29], and CNN-Transformer[30] networks are provided in the appendix. The detailed network structures for the models used in this study are provided in Table 2.

Table 2. Network structures of the methods compared.

| Models | Modules | Settings |
|---|---|---|
| CNN | Module 1 | Conv1D(1, 16, 3), BN, ReLU |

|  | Module 2 | Conv1D(16, 32, 3), BN, ReLU, AdaptiveMaxPool1D(10) |
|---|---|---|
|  | Module 3 | Linear(320, $n_c$) |
| CNN-LSTM | Module 1 | Conv1D(1, 32, 7), BN, ReLU |
|  | Module 2 | Conv1D(32, 64, 5), BN, ReLU, MaxPool1D(2) |
|  | Module 3 | Conv1D(64, 128, 3), BN, ReLU |
|  | Module 4 | Conv1D(128, 256, 3), BN, ReLU, AdaptiveMaxPool1D(1) |
|  | Module 5 | LSTM(256, 100, 3, dropout=0.5) |
|  | Module 6 | Linear(200, 512), BN, ReLU, Dropout(0.5) |
|  | Module 7 | Linear(512, 256), BN, ReLU, Dropout(0.5) |
|  | Module 8 | Linear(256, $n_c$) |
| CNN-GRU | Module 1 | Conv1D(1, 16, 3), BN, ReLU |
|  | Module 2 | Conv1D(16, 32, 3), BN, ReLU, AdaptiveMaxPool1D(10) |
|  | Module 3 | GRU(32, 128, 2, bidirectional=True) |
|  | Module 4 | Linear(256, $n_c$) |
| CNN-Transformer | Module 1 | Conv1D(1, 16, 3), BN, ReLU |
|  | Module 2 | Conv1D(16, 32, 3), BN, ReLU, AdaptiveMaxPool1D(10) |
|  | Module 3 | TransformerEncoder(32, 4, 128, 2) |
|  | Module 4 | Linear(32, $n_c$) |
| ResNet18 | Module 1 | Conv1D(1, 64, 7), BN, ReLU, MaxPool1D(3) |
|  | Module 2 | BasicBlock(64, 64)×2 |
|  | Module 3 | BasicBlock(64, 128, stride=2)×2 |
|  | Module 4 | BasicBlock(128, 256, stride=2)×2 |
|  | Module 5 | BasicBlock(256, 512, stride=2)×2 |
|  | Module 6 | AdaptiveAvgPool1D(1) |
|  | Module 7 | Linear(512, 256), ReLU, Dropout |
|  | Module 8 | Linear(256, $n_c$) |

The results in Table 3, Figure 9 and Figure 10 highlight how the proposed method provides a stable and efficient solution across diverse datasets. Unlike traditional approaches that either load a single source (single channel loading) or use multiple sources simultaneously (parallel channel loading), selective embedding switches between different data sources, enabling the benefits of multi-source information usually provided in parallel channels, while retaining the advantages of single-channel simplicity and efficiency. This balance improves generalization, avoids overfitting, and reduces computational demands.

While the traditional parallel loading method occasionally outperforms the proposed method with slightly higher accuracy (less than 3% difference for datasets 1, 5, and 6 when using a simple CNN), it comes at a significant computational cost. For instance, when using the CNN with dataset 5, parallel loading provides an accuracy of 98.22% after 110 seconds while selective embedding provides a very good accuracy of 95.31% after only 20 seconds. Nonetheless, selective embedding provides higher accuracies for most cases when compared to other data input methods, but at significantly reduced model training time (see appendix). This is especially true for more complex datasets, as can be seen with datasets 2, 3, and 4 in Table 3, Figure 9 and Figure 10, where selective embedding outperforms traditional data loading strategies. The key contribution of the proposed method lies in its efficiency and ability to stabilize all networks, significantly reducing sensitivity to noise. Across all datasets and algorithms tested, the method maintains over 80% accuracy (usually over 90%), even under challenging conditions like those in dataset 3, where traditional methods often fall short. Specifically, selective embedding achieves 84.82% (CNN) and 92.43% (CNN-Transformer) for dataset 3, a notable improvement over traditional data loading methods that remain below a 70% accuracy threshold.

Cases where the proposed method does not outperform traditional data loading strategies only occur when using simple datasets (e.g., datasets 1 – few classes and dataset 5 – few instances of each class). This reduced diversity in the data limits the advantage of alternating between sources. However, in more complex scenarios (e.g., dataset 3 or 4), where diversity in the data is high, selective embedding consistently and significantly outperforms other methods, demonstrating its strength for use in real-world applications where complexity and diversity are common.

In summary, the novelty of selective embedding lies in its ability to integrate multiple data sources in an alternating manner while reducing the total amount of data used, thereby mitigating the risk of overfitting and the high computational costs commonly seen in traditional data loading methods. Moreover, its ability to stabilize and improve the reliability of any algorithm or dataset makes it particularly suitable for implementation in industry. Selective embedding does not always outperform other data loading strategies when these strategies perform well already. However, this only occurs with simple datasets, and in these cases, the performance of the proposed method is only slightly less than traditional loading methods. Nonetheless, for difficult tasks, where traditional data loading strategies fail to provide adequate results, selective embedding maintains high performing results appropriate for industry. By providing stable and reproducible performance across various datasets and algorithms, selective embedding highlights a practical pathway towards achieving more efficient and human-like data processing in deep learning applications.

**Table 3.** Comparison of data loading methods when using a CNN and a CNN-Transformer

|  | Single Channel Data Loading (Traditional) | | Parallel Data Loading (Traditional) | | Selective Embedding (Proposed) | |
|---|---|---|---|---|---|---|
| **CNN** | Test Accuracy (%) | Time (s) | Test Accuracy (%) | Time (s) | Test Accuracy (%) | Time (s) |
| Dataset 1 | 93.37±1.92 | 10 | 94.53±2.11 | 65 | 92.21±1.16 | 10 |
| Dataset 2 | 82.37±2.67 | 10 | 86.72±2.04 | 65 | 95.64±0.31 | 10 |
| Dataset 3 | 50.38±3.02 | 25 | 55.40±3.64 | 200 | 84.82±2.72 | 25 |
| Dataset 4 | 80.49±4.05 | 15 | 84.65±4.84 | 135 | 94.21±5.69 | 15 |
| Dataset 5 | 97.06±0.65 | 20 | 98.22±1.23 | 110 | 95.31±1.58 | 20 |
| Dataset 6 | 98.70±1.61 | 60 | 99.93±0.17 | 110 | 97.64±4.95 | 60 |
| **CNN-Transformer** | Test Accuracy (%) | Time (s) | Test Accuracy (%) | Time (s) | Test Accuracy (%) | Time (s) |
| Dataset 1 | 77.54±4.00 | 20 | 79.21±3.64 | 190 | 80.62±2.07 | 20 |
| Dataset 2 | 90.69±5.57 | 20 | 92.74±4.53 | 270 | 95.95±1.00 | 20 |
| Dataset 3 | 64.54±8.65 | 40 | 67.41±7.51 | 460 | 92.43±4.16 | 40 |
| Dataset 4 | 78.19±4.42 | 35 | 82.54±4.05 | 280 | 96.55±1.50 | 35 |
| Dataset 5 | 99.44±0.47 | 35 | 99.48±0.78 | 225 | 99.14±1.00 | 35 |
| Dataset 6 | 99.36±0.66 | 120 | 99.64±0.54 | 140 | 98.64±2.67 | 120 |

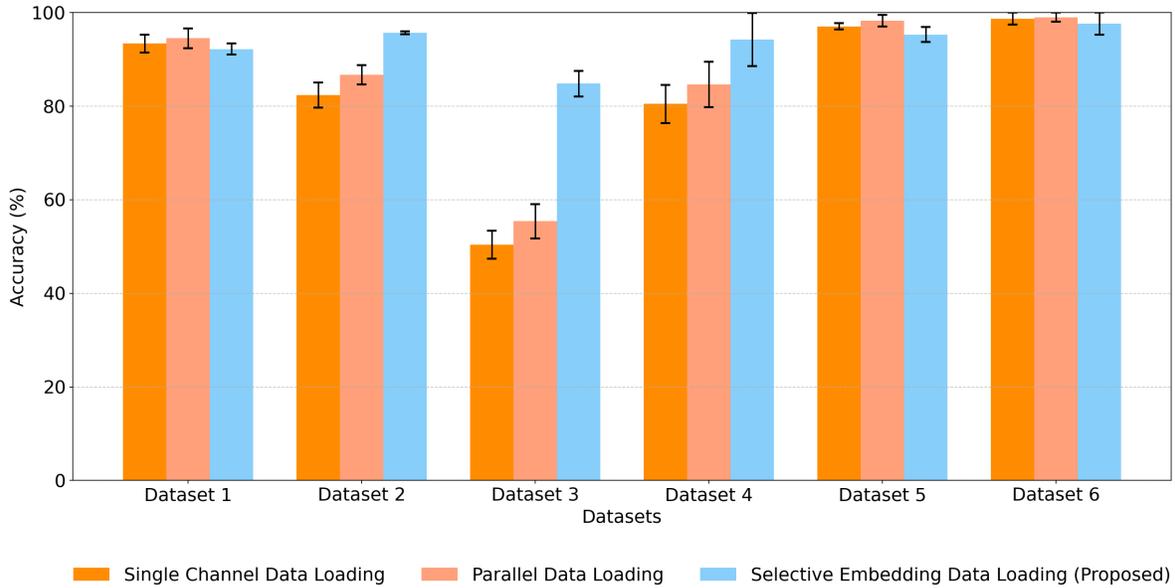

**Figure 9.** Test accuracy distribution for data loading methods for all datasets using a simple CNN architecture

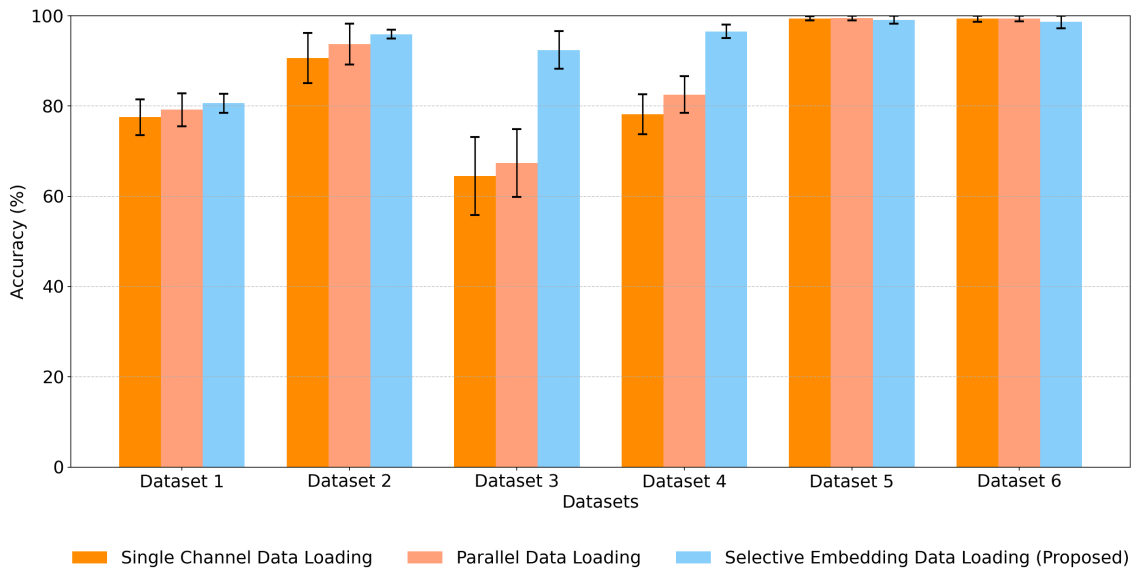

**Figure 10.** Test accuracy distribution for data loading methods for all datasets using a CNN-Transformer

## 4. Conclusion

Selective embedding provides a practical and efficient approach to improving model performance and robustness while reducing computational costs. By leveraging the richness of information obtained from different sources, this data loading method ensures consistent and reliable accuracy across datasets and algorithms while requiring only a portion of the original data, making it both resource-efficient and scalable. Unlike traditional data loading methods, which often lead to inconsistent accuracy results or increased computational demands, the proposed approach maintains stability and reliability across nonstationary conditions and dissimilar domains.

This method is particularly valuable for real-world applications involving time-series data, such as heavy machinery, marine, agriculture, medical, and railway vehicles, where data diversity and computational efficiency are critical. Its ability to retain high accuracy while optimizing data usage makes it suitable for deployment in machine learning systems that require fast processing and adaptation to complex environments. By improving data efficiency without compromising accuracy, the proposed method contributes to making machine learning more practical and accessible for a wide range of industries.

**Credit Author Statement**

**Mert Sehri:** Conceptualization, Methodology, Software, Validation, Formal Analysis, Investigation, Data Curation, Writing- Original Draft, Writing- Review & Editing, Visualization

**Zehui Hua:** Visualization, Writing- Original Draft **Francisco de Assis Boldt:** Writing- Review & Editing **Patrick Dumond:** Writing- Review & Editing, Supervision

**Appendix- Additional Results for Selective Embedding**

To further validate the performance of selective embedding, this section provides extended test results using five different deep learning architectures: CNN, RNN, GRU, CNN-Transformer, and ResNet18. Each model was evaluated across six benchmark datasets using two data loading strategies: multi-channel, parallel data loading and single-channel, multi-source data loading (selective embedding).

To visually compare the performance across datasets and architectures, Figure 11 and Figure 12, present the test accuracy distributions using radar plots for each respective model.

Across a majority of the cases, selective embedding consistently achieves higher test accuracy and lower variance, particularly on datasets 2, 3, and 4. These datasets are known to be more complex due to a higher diversity in the data. For these cases, selective embedding proves to be particularly effective, especially when paired with transformer-based architectures.

Importantly, even for simpler datasets (e.g., dataset 1 or 5), selective embedding performs well when compared to traditional data loading methods at a fraction of the computational cost. This highlights the method's adaptability, and scalability, making it suitable for both complex and straightforward machine learning tasks involving multi-source data.

These results reinforce the practical benefits of selective embedding for real-world machine learning tasks involving diverse sources of data, complex operating environments, and time-series signals. The method maintains high accuracy across architectures and datasets while reducing computational time or requirements.

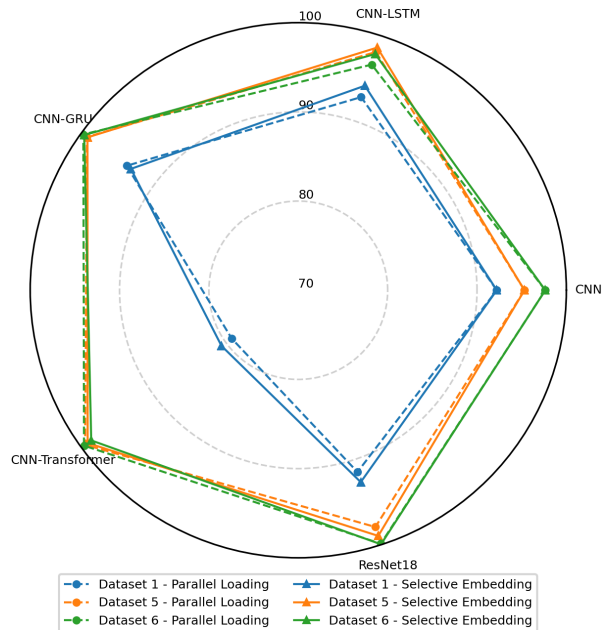

**Figure 11.** Radar plot using datasets 1, 5, and 6 for different machine learning algorithms

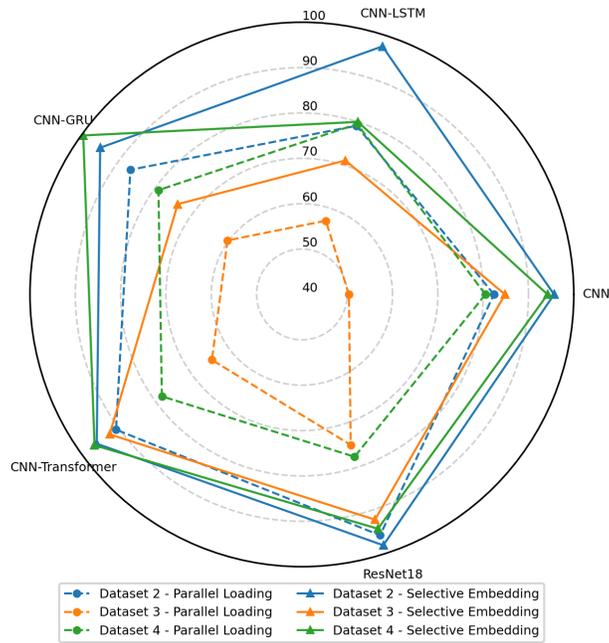

**Figure 12.** Radar plot using datasets 2, 3, and 4 for different machine learning algorithms